\documentclass[
	a4paper, 
	10pt, 
]{LTJournalArticle}

\usepackage{parskip}
\usepackage{multirow}
\usepackage{float} 

\usepackage{fix-cm}
\usepackage[T1]{fontenc}
\usepackage[fontsize=10.6pt]{fontsize}

\runninghead{Hawley \& Tackett, ``Operational Latent Spaces''} 

\footertext{\textit{Preprint. Accepted for 2024 AES International Symposium on AI and the Musician}} 

\usepackage{graphicx}
\graphicspath{{./}{figures/}{figs/}}

\usepackage{microtype}
\usepackage{amssymb}
\usepackage{subcaption}
\usepackage{overpic}

\usepackage{sectsty}
\sectionfont{\fontsize{13}{15}\selectfont}

\usepackage{algorithm}

\newcommand{\proj}{$h$}
\newcommand{\trans}{$T$}

\title{Operational Latent Spaces} 

\author{%
	Scott H. Hawley\textsuperscript{1,2}
 \thanks{Corresponding author: \href{mailto:scott.hawley@belmont.edu}{scott.hawley@belmont.edu} \textbf{Preprint:} \today},
         Austin R. Tackett\textsuperscript{1}
}

\date{\normalsize\textsuperscript{\textbf{1}}Belmont University, Nashville, TN, USA\\ \textsuperscript{\textbf{2}}Hyperstate AI}

\renewcommand{\maketitlehookd}{%
	\begin{abstract}
		\noindent We investigate the construction of latent spaces through self-supervised learning to 
support semantically meaningful operations. 
Analogous to operational amplifiers, these "operational latent spaces" (OpLaS) not only demonstrate semantic structure such as clustering but also support common transformational operations with inherent semantic meaning. 
Some operational latent spaces are found to have arisen
``unintentionally" in the progress toward some (other) self-supervised learning objective, in which unintended but still useful properties are discovered among the relationships of points in the space. 
Other spaces may be constructed ``intentionally" by developers stipulating certain
kinds of clustering or transformations intended to produce the desired structure.
We focus on the intentional creation of operational latent spaces via self-supervised learning, including the introduction of rotation operators via a novel ``FiLMR'' layer, which can be used to enable ring-like symmetries found in some musical constructions. 
	\end{abstract}
}

\begin{document}

\maketitle



\section{Introduction}

Self-supervised learning has emerged as a powerful tool for uncovering latent representations within data. These latent spaces, often high-dimensional, capture the underlying structure of the data in a way that can be surprisingly meaningful. Notably, some latent spaces exhibit the remarkable property of supporting transformations that correspond to real-world manipulations with semantic interpretations. These transformations can often be expressed as translations or scaling within the space, allowing for intuitive control over the data.

Examples of this can be found in natural language processing, where algebraic manipulations of word vectors can encode complex relationships. 
For instance, the well-known equation "king" - "man" + "woman" = "queen" from Word2Vec \cite{word2vec} exemplifies how these vectors capture semantic relationships. Similarly, it has been observed that vectors representing countries and their capitals often lie along parallel lines within the latent space, reflecting a clear geometric relationship. These geometric relationships were not necessarily intended but were later explained as having arisen due to the use of matrix factorization in the optimization objective \cite{glove}. Matrix factorization was then explicitly used as an objective to encourage semantic geometric structures in subsequent models. Matrix factorization is employed in style transfer systems \cite{styletransfer1, styletransfer_fonts}, as factorization is one mechanism for disentangling representations \cite{kim2019disentangling}.


Disentangled representation learning, where the latent space factors correspond to independent aspects of the data, is another promising approach for achieving controllable music generation \cite{fewshotvoice, MMST_tony}. Recent work has also explored leveraging relative positioning within the latent space to control audio effects \cite{moschella2023relative}.
 In the image domain, StyleGAN and StyleGAN2 were built upon the premise of 
disentangling controls of image generation. It was later
discovered \cite{ganspace} that many other types of possible
controls are ``latent'' within StyleGAN beyond what it was
originally intended for, including controls for 
subjective criteria such as ``fluffiness."
The potential to unlock new semantic controls within audio,
using the latent space of {\it pre-trained} audio models has 
received some preliminary attention \cite{hawley_steinmetz_2023} but the spaces were found to be highly nonlinear, even for linear audio transformations high as high-pass or low-pass filtering.
Thus, we may wish to modify the existing latent space of the pretrained model to support the operations we wish to perform, using projective methods such as SimCLR \cite{simclr} or
 VICReg \cite{VICReg}, which have proved to be powerful tools
 for self-supervised representation learning. 


This paper investigates the potential for self-supervised learning applied to the latent spaces of pretrained audio encoding models to create interpretable latent spaces that empower music producers with fine-grained control over generative models. We present our approach, evaluate its effectiveness, and discuss the implications for fostering creative expression within music production. We consider the effects of enforcing algebraic structures onto the geometry of the latent space, applied through metric learning losses in self-supervised ways.  This work bears similarity with some work on ``task arithmetic'' \cite{task_arith1, task_arith2}, and the desire to exploit symmetries \cite{liu2023learning} to achieve musically relevant data transformations, yet offers a different set of tasks and mechanisms.

In Section \ref{sec:aa}, we seek to recover a vector space for music mixing in the latent domain.  In Section \ref{sec:rot}, we go beyond translations and scaling to include rotations among the operations used to provide semantic relationships between data points. 
We provide supplemental materials and code via a companion website\footnote{Demo \& code: \href
{https://drscotthawley.github.io/oplas}{https://drscotthawley.github.io/oplas}}.

\section{Example 1:\  Mixing in Latent Space}\label{sec:aa}

\begin{figure}
    \centering
    \includegraphics[width=\columnwidth]{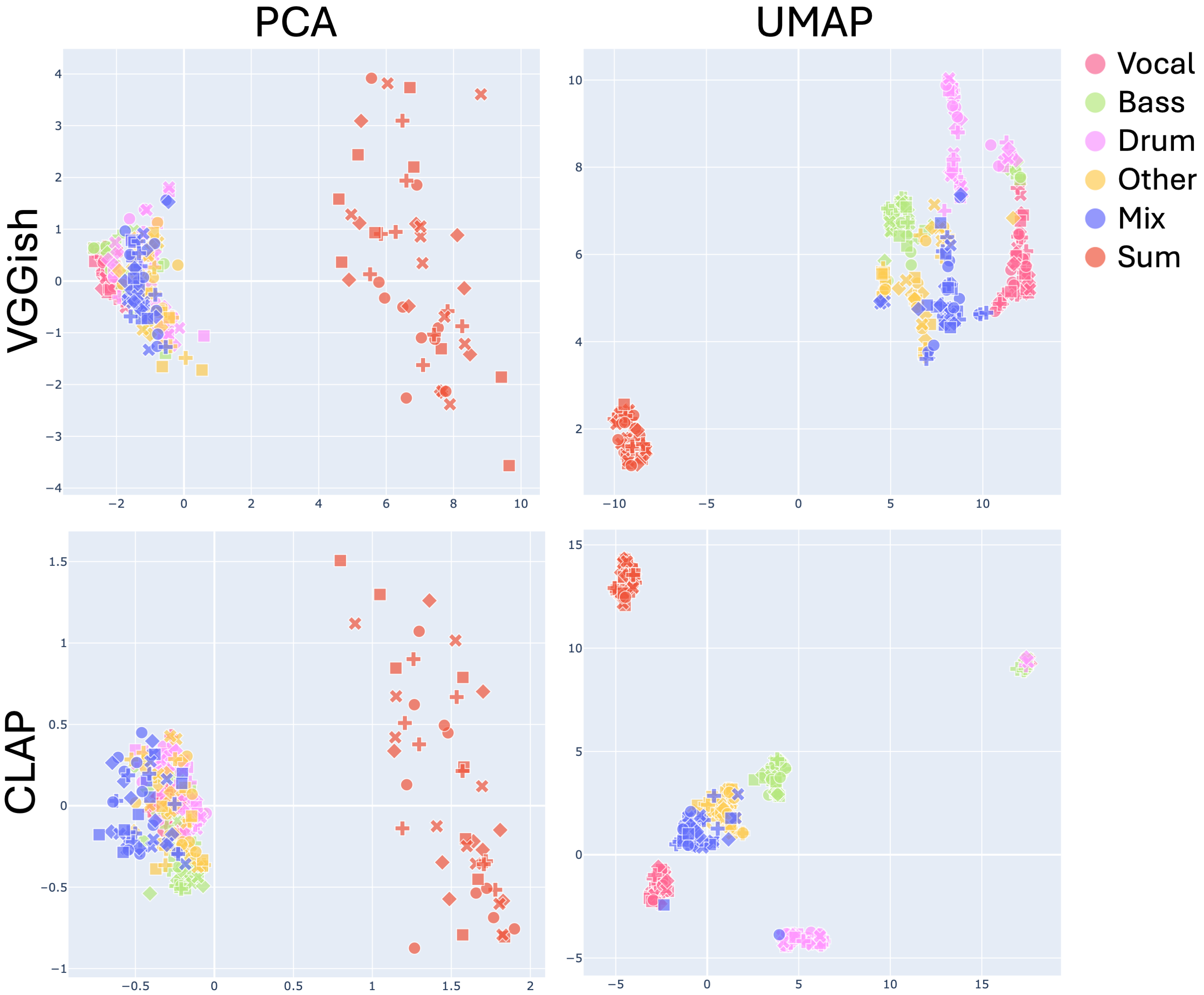}
    \captionsetup{format=plain}
    \caption{Encoded stems and mixes from the MUSDB18 \cite{MUSDB18} audio dataset using the VGGish (top row) and CLAP (bottom row) pretrained encoding models, visualized using PCA (left column) and UMAP (right column).
    We see that while different stems encode to similar locations, their sums (brown markers) are far from the mix encodings (purple markers), illustrating the nonlinearity of these encoding models.
    }
    \label{fig:austin-dots}
\end{figure}

In typical linear mixing environments such as in the time or spectral (i.e. Fourier) domains, the ``mix'' is simply a weighted sum of the component musical parts.  Neural network systems for audio processing, however, typically incorporate nonlinear transformations which may prevent the sums of neural activations from accurately representing the audio mix.  
How ``nonlinear" are typical neural audio embeddings?  In Figure \ref{fig:austin-dots}, we take various stem components from the MUSDB18 dataset, sum them, and encode them into latent space using VGGish \cite{vggish} and CLAP \cite{laionclap2023}.

We consider a ``toy model'' of points in two dimensions, generating (neural) embeddings via some example nonlinear process, and wish to accomplish the following: {\it find a ``projector'' mapping $h$ from the embedding domain into another domain in which the sum of the embeddings lies arbitrarily close to the embedding of the full musical mix}.  We could also require that $h$ possess an (approximate) inverse $\widehat{h^{-1}}$ which would allow the projective space to comprise a {\em ``latent plugin''} for the pretrained given model $f$.

\begin{figure*}[ht!]
\centering
\vspace*{0cm}

\begin{minipage}[t]{\textwidth} 
    \hspace{1.7cm}\begin{overpic}[width=.9\textwidth]{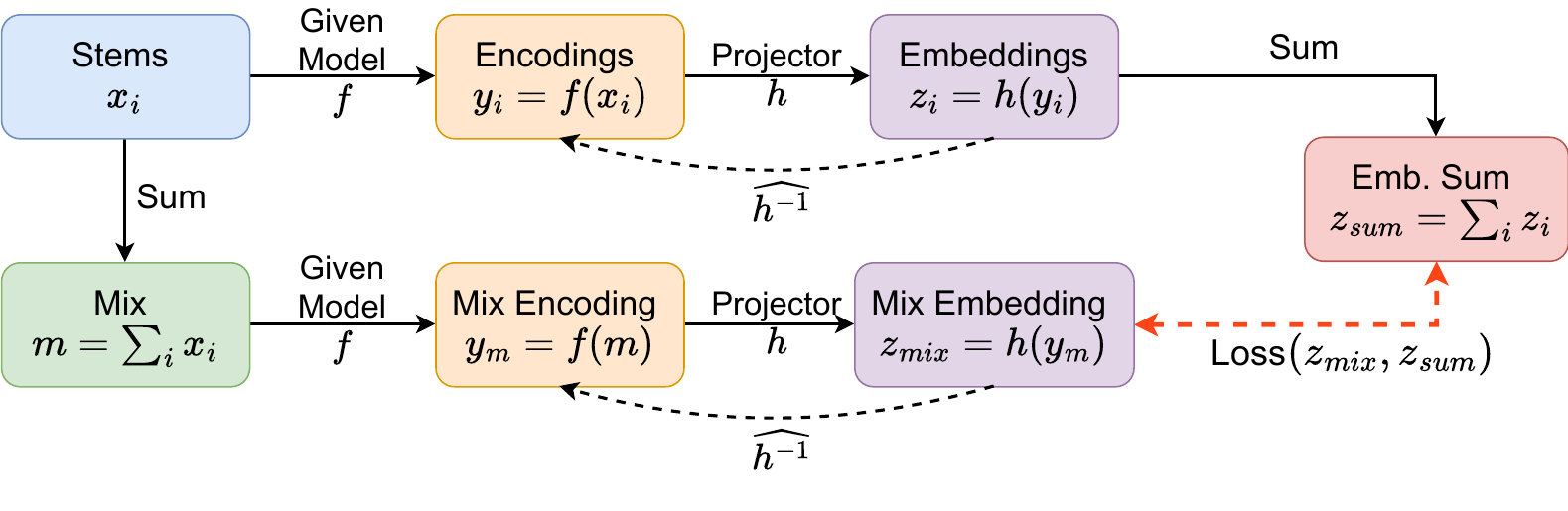}
        \makebox(-10,60){{a)}}
    \end{overpic}
\end{minipage}

\begin{minipage}[t]{\textwidth} 
    \hspace{1.7cm}\begin{overpic}[width=.9\textwidth]{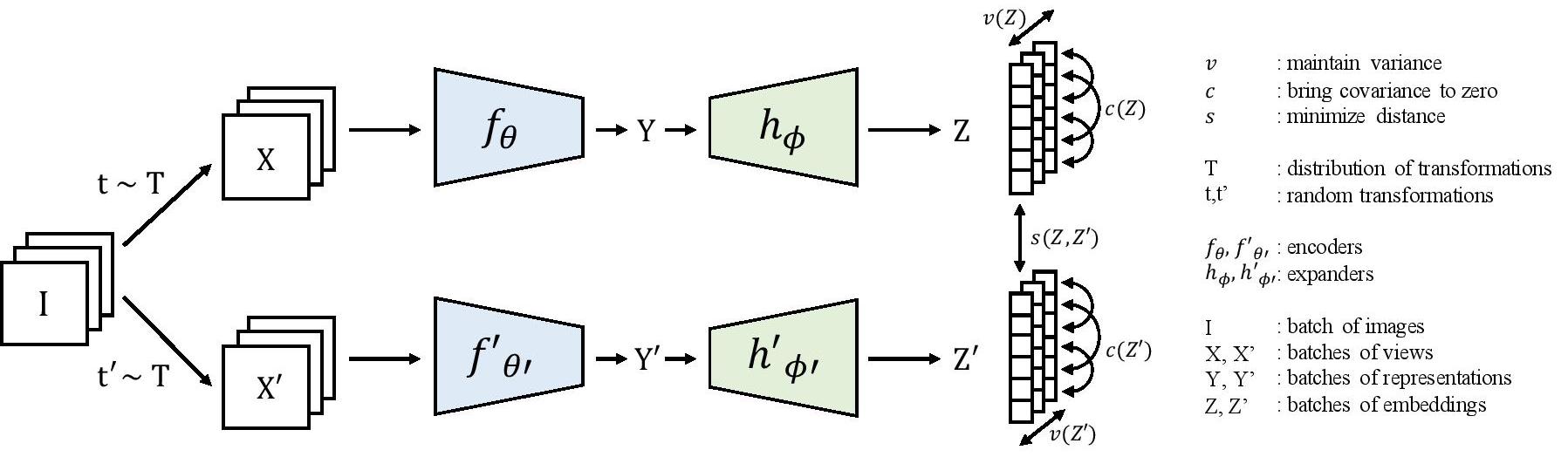}
        \makebox(-10,47){{b)}}
    \end{overpic}
\end{minipage}
\vspace{0cm}

\begin{minipage}[t]{\textwidth} 
    \hspace{1cm}\begin{overpic}[width=.95\textwidth]{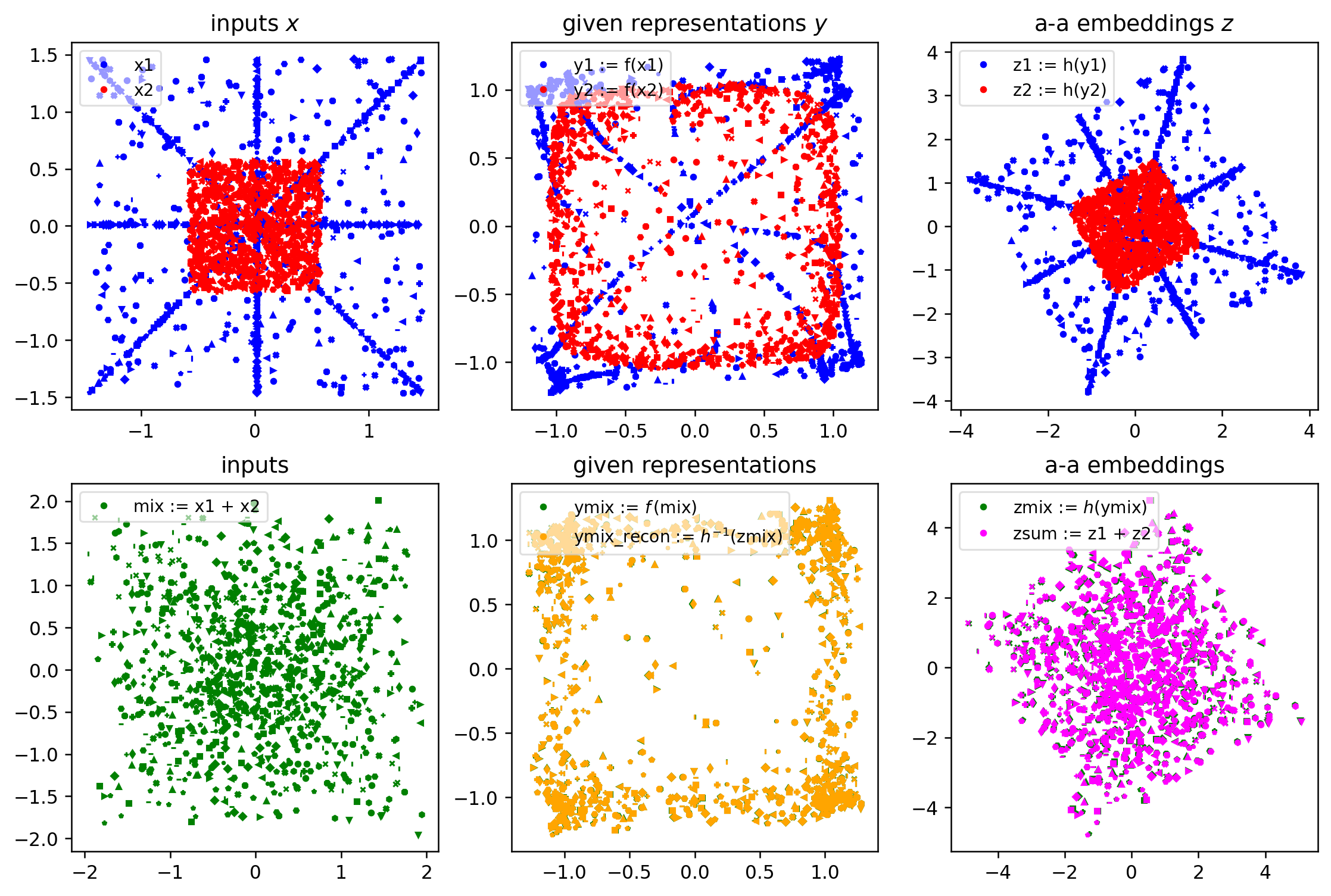}
        \put(-2,62){{c)}}
    \end{overpic}
\end{minipage}
\captionsetup{singlelinecheck=false}
\captionsetup{format=plain}
\caption{
 Mixing with embeddings.  a) Flowchart of the algorithm, inspired by a similar flowchart from the VICReg paper \cite{VICReg} shown in b) for comparison.  c) Implementation using two classes of 2-D "dots'' as proxies for audio stems. The sum of the stems $x_i$ appears in the bottom left in green as the ``mix''.  In the middle column, we apply some nonlinear twisting and leveling to the ``dots'' in the left column.  In the bottom right, the sums of the embeddings (purple shapes) lie right on top of the embeddings of the mixes (green shapes).
  Finally, the yellow dots in the bottom middle covering the green dots confirm that we have learned an invertible mapping. 
}
\label{fig:aa_flow}
\end{figure*}

Figure \ref{fig:aa_flow} illustrates a schematic for the neural network architecture used, similar to the setup of VICReg \cite{VICReg} yet applied to a new purpose. 


\begin{figure}[ht]
\vspace{-.1cm}
\centering
\includegraphics[width=.9\columnwidth]{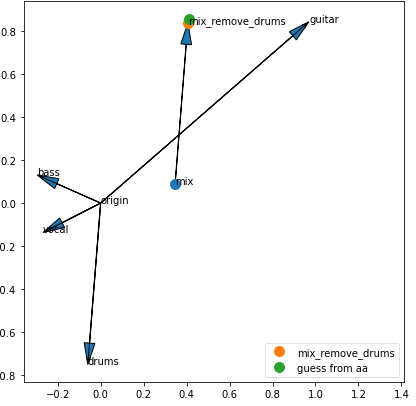}
\captionsetup{format=plain}
\caption{Mixing in latent space: subtracting the ``drums'' vector.  
Here, the signals denoted by ``vocal,'' ``bass,'' ``drums'' and their time-domain sum ``mix`` are first embedded in a space $Y$ and then projected into $Z$.  We then compare the projected vectors for the mix without the drums (in the time domain) shown in orange with the ``audio algebra'' result of subtracting the vector for ``drums`` from the ``mix`` vector. We see that these are very close to each other in the projected space $Z$. 
}
\label{fig:latent space}
\end{figure}
\vspace{-.7cm}


\vspace{6mm}
This preliminary toy study suggests that semantic audio transformations in latent space may be constructed explicitly using self-supervised representation learning techniques.  Similar studies using real audio encoded via VGGish and/or CLAP models are currently underway but are incomplete at the time of manuscript submission.

\section{Example 2:\ Enabling Rotations}\label{sec:rot}

Beyond the FiLM layers \cite{film}, which learn abelian transformations, we can try adding rotations, which may lead to additional (and perhaps powerful) semantic operations.  We refer to such layers as ``FiLMR'' layers with the ``R'' denoting the inclusion of rotation transformations.   To illustrate the potential utility for rotations, we pose a sample problem in two dimensions (the ``Stargate Problem'', below) to compare a network using square matrices instead of FiLMR layers.

Beyond 2 dimensions, arbitrary rotations in $n$ dimensions incur a ``curse of dimensionality'' since their symmetry group has a ``triangular number`` of $(n^2 - n)/2$  degrees of freedom. Restricting our attention to ``simple rotations'' in a 2-dimensional subspace, we still retain functionality. The algorithm of Aguilera and Aguila \cite{Aguilera2004GeneralNR} provides a way to construct such a rotation operator $M$ iteratively using the plane of 2 $n$-dimensional vectors $\vec{u}$ and $\vec{v}$, rotating arbitrary vectors $\vec{x}$ by twice the angular separation of $\vec{u}$ and $\vec{v}$.   Algorithm 1 shows an outline of a FiLMR layer's operation.

\begin{algorithm}[b]

\caption{FiLMR Layer in $n$ dimensions}\label{alg:filmr}
\fontsize{10.5}{12}\selectfont
 Trainable Parameters: $\gamma, \beta \sim \mathcal{N};$ 
$\ \ \vec{u}, \vec{v} \sim \mathcal{N}(\mathbf{0}, I_n)$

\vspace{-3mm}
\hrulefill{\vspace*{-1mm}} 

Forward method: \\
\hspace*{.1cm}Compute rotation matrix $M(\vec{u}, \vec{v}) \in \mathbb{R}^{n \times n}$ via \cite{Aguilera2004GeneralNR} \\
\hspace*{.1cm}Given input $\vec{x} \in \mathbb{R}^n$, transform 
 $\vec{x} \leftarrow (\gamma \vec{x} + \beta)  M$
\end{algorithm}

\subsection{The "Stargate Problem"}\label{subsec:starget}

As an example toy problem, we imagine giving the model the task of creating a latent space supporting a simple operation: given a data element (i.e., data point) advance to the next point, with a wrap-around boundary condition such that if the point in question is the last element in the sequence, the operation will map to the first point in the sequence.  This is a classic specification of a ``ring'' symmetry group.  Such rings occur in many fields, but especially so in musical contexts such as the basic modulo-12 arithmetic of musical keys, the Circle of Fifths, and the ``matrix'' of John Coltrane \cite{coltrane1960giantsteps}.  Our sample problem is very simple, but we could extend this by imagining tasks such as: What if we wanted to embed the Coltrane Matrix in a latent space and learn the transformations for Coltrane’s processes?

\begin{figure*}[h!]
    \centering
    
    \begin{minipage}[t]{\textwidth} 
        \centering
        \hspace{1.7cm}\begin{overpic}[width=.85\textwidth]{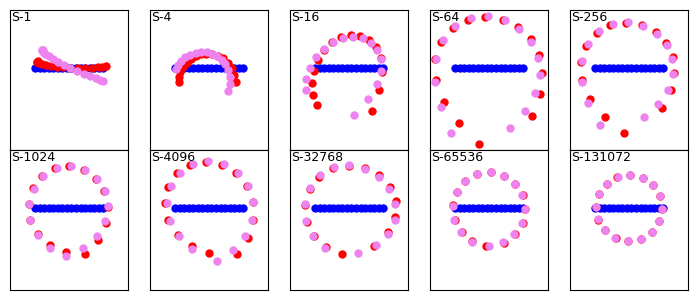}
            \makebox(-4,72){{a)}}
        \end{overpic}
    \end{minipage}
    
    
    \begin{minipage}[t]{\textwidth} 
        \centering
        \hspace{1.7cm}\begin{overpic}[width=.85\textwidth]{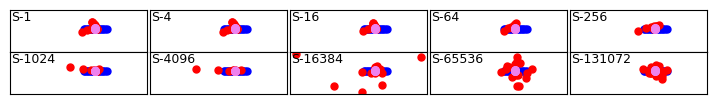}
            \put(-4,12){{b)}}
        \end{overpic}
    \end{minipage}

    \caption{a) Progress of the Stargate Problem using FiLMR layer.  "S-" in the top left of each pane indicates the training step number.
    b) In contrast, evolution using a learned square orthogonal matrix. While such a solution should exist in theory, the neural network fails to learn the appropriate transformations, perhaps due to dynamic instability.
    See Figure \ref{fig:stargate-end} for a zoomed view of the final simulation states.
    }
    \label{fig:stargate-frames}
    \vspace{-.3cm}
\end{figure*}

\begin{figure}[h!]
\centering
\begin{minipage}[t]{\columnwidth} 
        \centering
        \hspace{1cm}\begin{overpic}[width=.78\columnwidth]{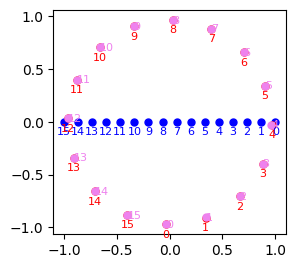}
            \makebox(-4,165){{a)}}
        \end{overpic}
    \end{minipage}
     \begin{minipage}[t]{\columnwidth} 
        \centering
        \hspace{1cm}\begin{overpic}[width=.78\columnwidth]{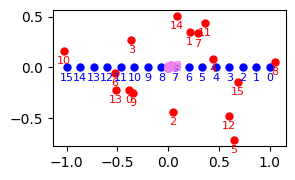}
            \put(-4,50){{b)}}
        \end{overpic}
    \end{minipage}
\centering
\captionsetup{format=plain}
    \caption{a): ``Final'' successful state of model trying the Stargate Problem via a FiLMR layer.  The red and pink colors and numbers are intended to show points lining up on top of their ``targets,'' {\em i.e.,} the next points in the sequence. b): Unsuccessful result of trying to use a learned orthogonal square matrix. 
    }
    \label{fig:stargate-end}
    \vspace{-.5cm}
\end{figure}

Formally, this means, given some initial data space $Y$, the model learns a projection \proj to a new space $Z$ such that for points $z_i \in Z$, the model is also able to learn a transformation \trans such that $T(z_i) = z_{i+1}$.
We refer to this learning task as the ``Stargate Problem'' because watching the system try to ``lock in'' while learning the ring structure is somewhat reminiscent of ``Stargate'' movie and TV shows, in which getting the stargate's chevron-shaped elements to ``lock'' was a prerequisite for interstellar travel.

We start with points $y_i \in Y$ that lie along a horizontal line shown in blue in Figure \ref{fig:stargate-frames}.
Even though we may ``know`` that the correct pair of functions \proj, \trans to learn are those in which points $z_i$ in the new space $Z$ form a circle, and that \trans should simply be a rotation of $2\pi/N$ (if $N$ is the number of data points)  One {\em could} apply such goals in the form of supervised learning which would make this problem nearly trivial. Instead, we make the model minimize the objective
\begin{equation}
  \left[ T(z_i) - z_{i+1} \right]^2 
\end{equation}
where $z_i = h(y_i)$. Note also that the points $y_i$ are imagined as encodings of time-domain audio $x_i$, encoding via $y_i = f(x_i)$.


 In theory, learning a square matrix for both \proj and \trans could produce the desired projection and transformation properties, respectively.  In practice, however, we find trying to learn a full matrix doesn't work, i.e. none of the many attempts we tried ever resulted in the desired structure.  Instead, the square-matrix solutions tend to extend the points $z_i$ along a line.  Figure \ref{fig:stargate-frames} illustrates the progress of training, with final states shown in Figure \ref{fig:stargate-end}.

\begin{figure}[h!]
    \centering
    \includegraphics[width=0.85\columnwidth]{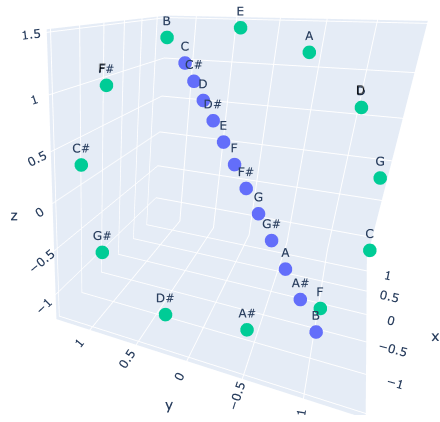}
    \captionsetup{format=plain}
    \caption{PCA plot after extending the Stargate Problem to 64 dimensions and converting the sequence of notes to the musical Circle of 5ths. The inputs lie along the diagonal line of 1's (1,1,1,...), and the system learns a rotation operator to bring them into a ring.  A separate process learns to rearrange the notes according to the Circle of 5ths.}
    \label{fig:co5}
\end{figure}

Extending beyond 2 dimensions to $n$ dimensions, we 
make use of the Aguilera-Perez Algorithm \cite{Aguilera2004GeneralNR} to construct a $n$-dimensional rotation matrix $M$ from two learned 
$n$-dimensional vectors $\vec{u}, \vec{v}.$ 
The action of $M$ will be to rotate in the plane of 
$\vec{u}$ and $\vec{v}$ by an angle double that of their separation.  Thus, rather than needing to learn a ``triangular number'' of O($n^2$) parameters, the 
system only needs to learn $2n$ parameters beyond the initial scale and translation of the FiLM layer. 
The $n$-dimensional FiLMR layer is outlined 
in Algorithm \ref{alg:filmr}.  In Figure \ref{fig:co5} we show the result of learning a ring structure in 64 dimensions, along with a rotation of half the elements to form the musical ``Circle of Fifths.''

\section{Summary}

We have shown two examples of constructing ``operational latent spaces (OpLaS)," via self-supervised learning, taking the encodings from larger pretrained models and projecting them to spaces that support a desired (learned) transformation such as summation or rotation. These systems show potential for enabling ``latent plugins'' for larger pretrained models which by default may not support the desired transformations. 
The pointwise actions of the loss functions in these systems are reminiscent of inter-particle forces in 
physics, which typically arise via some symmetry such as energy conservation \cite{dawid2023introduction}.  This suggests that physical symmetries may yield a fruitful set of transformations for 
semantic musical operations, as is suggested by recent work by Liu et al \cite{liu2023learning}.
This paper serves as a preliminary feasibility study using ``points'' in space as proxies for audio stems and their encodings. Future work should include applying the techniques from this study to high-dimensional encodings of real audio.

\section{Acknowledgements}
The authors wish to thank Brad Schleben and Jonathan Rowden for stimulating discussions. Computing support was provided by equipment grants from Dell and Belmont University. The work for the ``Mixing in Latent Space'' section was largely completed in 2022 while S.H.H. was employed by Harmonai/Stability AI \cite{audio-algebra-talk}. The abbreviation ``OpLaS'' was suggested by Pablo Rivas. 

\printbibliography

\end{document}